\documentclass[12pt,a4paper]{article}

\usepackage{enumitem}
\usepackage{amsmath}
\usepackage{amsthm}
\usepackage{amssymb}
\usepackage{graphicx}
\usepackage{xcolor}

\usepackage{hyperref}
\usepackage{caption}
\usepackage{setspace}
\usepackage{mdframed}
\usepackage{authblk}

\usepackage{times}

\theoremstyle{plain}

\theoremstyle{remark}
\newtheorem{rem}{Remark}

\begin{document}
% Part 1

\global\long\def\apa{\alpha}%
 
\global\long\def\ba{\beta}%
 
\global\long\def\ga{\gamma}%
 
\global\long\def\Ga{\Gamma}%

\global\long\def\Sa{\Sigma}%
 
\global\long\def\da{\delta}%
 
\global\long\def\Da{\Delta}%
 
\global\long\def\ta{\theta}%
 
\global\long\def\Ta{\Theta}%
 
\global\long\def\la{\lambda}%
 
\global\long\def\La{\Lambda}%
 
\global\long\def\oa{\omega}%
 
\global\long\def\Oa{\Omega}%
 
\global\long\def\vn{\varepsilon}%
 
\global\long\def\vi{\varphi}%
 
\global\long\def\sa{\sigma}%
 
\global\long\def\ka{\kappa}%
 
% Part 2

\global\long\def\pl{\partial}%
 
\global\long\def\na{\nabla}%
 
\global\long\def\sto{\rightsquigarrow}%
 
\global\long\def\wto{\rightharpoonup}%
 
\global\long\def\To{\Rightarrow}%
 
\global\long\def\Lrto{\Leftrightarrow}%
 
\global\long\def\fc#1#2{\frac{#1}{#2}}%
 
\global\long\def\st#1{\sqrt{#1}}%
 
\global\long\def\ol#1{\overline{#1}}%
 
\global\long\def\ul#1{\underline{#1}}%
 
\global\long\def\wt#1{\widetilde{#1}}%
 
\global\long\def\wh#1{\widehat{#1}}%
 
% Part 3

\global\long\def\mf#1{\mathbf{#1}}%
 
\global\long\def\mb#1{\mathbb{#1}}%
 
\global\long\def\ml#1{\mathcal{#1}}%
 
\global\long\def\mk#1{\mathfrak{#1}}%
 
\global\long\def\mm#1{\mathrm{#1}}%
 
\global\long\def\mr#1{\mathscr{#1}}%
 
\global\long\def\bs#1{\boldsymbol{#1}}%

\fontsize{12}{16pt plus.4pt minus.3pt}\selectfont
%\setstretch{1.2}

\title{A new locally linear embedding scheme in light of Hessian eigenmap}
%\author{Liren Lin\footnote{lirenlin2017@gmail.com}\quad and\quad 
%Chih-Wei Chen\footnote{chencw@math.nsysu.edu.tw}\\[5pt]
%Department of Applied Mathematics, \\ National Sun Yat-sen University, Taiwan
%}
\author[]{Liren Lin\footnote{lirenlin2017@gmail.com}\ \ }
\author[]{\ Chih-Wei Chen\footnote{chencw@math.nsysu.edu.tw}}
\affil{Department of Applied Mathematics, National Sun Yat-sen University, Taiwan}
%\affil[2]{Department of Mathematics, National Taiwan University, Taiwan}
\date{}

\maketitle

\begin{abstract}
We provide a new interpretation of Hessian locally linear embedding
(HLLE), revealing that it is essentially a variant way to implement the 
same idea of locally linear embedding (LLE). Based on the new
interpretation, a substantial simplification can be made, in which the 
idea of ``Hessian'' is replaced by rather arbitrary weights. 
Moreover, we show by numerical examples that HLLE may 
produce projection-like results when the dimension of the 
target space is larger than that of the data manifold, and hence 
one further modification concerning the manifold dimension is
suggested. Combining all the observations, we finally achieve
a new LLE-type method, which is called tangential LLE (TLLE).
It is simpler and more robust than HLLE. 
\end{abstract}

\section{Introduction}

Let $\ml X=\{x_i\}_{i=1}^N$ be a collection of data points in 
some $\mb R^D$.
The goal of nonlinear dimensionality reduction 
(or manifold learning) is to find for $\ml X$
a representation $\ml Y=\{y_i\}_{i=1}^N$ in some lower 
dimensional $\mb R^d$, under the assumption that 
$\ml X$ lies on some unknown submanifold $\ml M$ in $\mb R^D$. 

Among the several existing manifold learning methods, 
Hessian eigenmap \cite{hlle},
also called Hessian locally linear embedding (HLLE), is one 
that exhibits prominent performance on the popular synthetic
data ``Swiss roll with a hole''. 
It can be regarded as a generalization of 
Laplacian eigenmap \cite{belkin2003laplacian}
in some respect or LLE \cite{lle} in another.
However, its procedure concerning the construction 
and minimization of ``Hessian'' is much more 
sophisticated. 

In this paper, we will provide a new interpretation of the 
mechanism behind HLLE,
revealing that what it really does follows the same idea as LLE:
Asking $\ml Y$ to satisfy the local linear relations 
for $\ml X$ as best as possible. The main differences lie in their 
ways of describing the local linear relations. 
Roughly speaking, HLLE only fits
local linear relations of the $d$-dimensional principal components,
and at the same time exploits multiple weights to do this job. 
Based on this understanding, we are able to make a substantial 
simplification, in which the idea of ``Hessian'' is totally
abandoned. Indeed, we will show that the
Hessian estimators in HLLE, which are specially designed matrices,
can be replaced by rather arbitrary weight matrices.

Moreover, we observe that when the dimension of the target space
$d$ is greater than that of the data manifold $\ml M$, 
a naive application of HLLE may result in a type of unwanted result, which 
looks like some direct projection of $\ml X$ onto $\mb R^d$.
Such ``projection patterns'' are also observed for LLE (see 
Section \ref{sec:llehlle}, or \cite{lin2021avoiding} for detailed discussion).
For HLLE, they do not appear in the embedding of the Swiss roll in the plane, 
for which $d=\dim\ml M=2$.
However, in general a manifold may not be able to be well embedded in a Euclidean 
space of the same dimension, and setting $d$ to be larger than 
$\dim\ml M$ might be more suitable. In such a situation, 
some modification has to be made to avoid the projection patterns. 
Combining the mentioned simplification 
and this modification, we finally achieve a new LLE-type method, which will be 
named tangential LLE (TLLE). 

\section{New interpretation of HLLE}\label{sec:hlle}

As in the introduction, in the following let $\ml X=\{x_{i}\}_{i=1}^{N}$ be a 
dataset in $\mb{R}^D$, which is supposed to lie on some 
unknown submanifold $\ml M$ (sometimes referred to as the data manifold),
and our goal is to find for $\ml X$ a representation
$\ml Y=\{y_{i}\}_{i=1}^{N}$ in some lower dimensional $\mb R^{d}$.

We first review the procedure of HLLE, given as \ref{H1} $\sim$ \ref{H4} below. The idea 
will be (partly) discussed right after.
\begin{enumerate}[align=left, label=(H\arabic{enumi})]
\item \label{H1} For each $i=1,\ldots,N$ let $\ml U_{i}=\{x_{i_{1}},\ldots,x_{i_{k}}\}\subset\ml X\setminus\{x_{i}\}$
be a $k$-nearest neighborhood of $x_{i}$, where 
\begin{align*}
k\ge 1+d+\fc{d(d+1)}2.
\end{align*}
For simplicity we consider $k$ to be a fixed number for all $i$.
\item \label{H2} Let 
\[
M^{(i)}=\begin{bmatrix}x_{i_{1}}-\ol x_{i} & \cdots & x_{i_{k}}-\ol x_{i}\end{bmatrix}
\in\mb{R}^{D\times k},
\]
where $\ol x_{i}=\fc 1k\sum_{j=1}^{k}x_{i_{j}}$. Find the 
singular value decomposition
\[
M^{(i)}=U^{(i)}\Sa^{(i)}V^{(i)T},
\]
where $U^{(i)}$ is a $D\times D$ orthogonal matrix whose columns will be 
denoted by $u_{1}^{(i)},\ldots,u_{D}^{(i)}$,
$\Sigma^{(i)}$ is a $D\times k$ diagonal matrix 
with diagonal entries $\sa_{1}^{(i)}\ge\cdots\ge\sa_{\min(D,k)}^{(i)}\ge 0$,
and $V^{(i)}$ is a $k\times k$ orthogonal matrix 
with columns $v_{1}^{(i)},\ldots,v_{k}^{(i)}$.
\item \label{H3}
Let $V_d^{(i)}=[v_{1}^{(i)}\ \cdots\ v_{d}^{(i)}]\in\mb{R}^{k\times d}$ 
be the matrix consisting of the first $d$ columns of 
$V^{(i)}$, and let 
$V_{d}^{(i)}\boxtimes V_{d}^{(i)}\in\mb{R}^{k\times d(d+1)/2}$ be the
matrix whose columns are all the vectors
$v_{s}^{(i)}v_{t}^{(i)}$, $1\le s\le t\le d$,
listed in any prescribed order. 
Here $ab$ for two (column) vectors
$a=(a_{1},\ldots,a_{k})$ and $b=(b_{1},\ldots,b_{k})$ denotes their 
componentwise product $(a_{1}b_{1},\ldots,a_{k}b_{k})$.
Let $\bs 1_k\in\mb{R}^k$ be the vector whose components are all equal to $1$,
and let 
\begin{equation*}
G^{(i)}=\begin{bmatrix}\bs 1_{k} & V_{d}^{(i)} & V_{d}^{(i)}\boxtimes V_{d}^{(i)}
\end{bmatrix},
\end{equation*}
which is a matrix of size $k\times(1+d+\fc{d(d+1)}2)$. Apply the
Gram-Schmidt process to the columns of $G^{(i)}$ to obtain a new
matrix with orthonormal columns, and let $H^{(i)}$ be the submatrix of this new matrix
consisting of the last $d(d+1)/2$ columns (the position of 
$V_{d}^{(i)}\boxtimes V_{d}^{(i)}$ in $G^{(i)}$). 
\item \label{H4} Set $\ml Y$ to be a solution to the minimization problem
\begin{equation*}
\underset{\{y_{i}\}_{i=1}^{N}\subset\mb R^{d}}{\mbox{argmin}}\sum_{i=1}^{N}\|Y_{i}H^{(i)}\|^{2}\quad\mbox{s.t.}\quad YY^{T}=I,%\label{eq:hmm}
\end{equation*}
where $Y=\left[y_{1}\ \cdots\ y_{N}\right]\in\mb R^{d\times N}$ and
$Y_{i}=\left[y_{i_{1}}\ \cdots\ y_{i_{k}}\right]\in\mb R^{d\times k}$.
Here $\|A\|$ for a matrix $A=(a_{ij})$ denotes the Frobenius norm
$\sqrt{\sum_{i,j}a_{ij}^2}$.
\end{enumerate}

Before giving our new interpretation of HLLE, 
we make some preliminary remarks and comments about its original idea.
First, what \ref{H2} does is to perform the principal component analysis for $\ml U_i$.
The $d$-dimensional principal
component of $x_{i_{j}}$ ($j=1,\ldots,k$), denoted by $\wt x_{i_{j}}$ later on, 
is the orthogonal projection of $x_{i_{j}}$ on
the $d$-dimensional hyperplane
\begin{equation}
\ml T_{i}:=\ol x_{i}+\mbox{span}(u_{1}^{(i)},\ldots,u_{d}^{(i)}).\label{eq:ti1}
\end{equation}
Accordingly, we will use $\wt{\ml U}_{i}$ to denote the set 
$\{\wt x_{i_1},\ldots,\wt x_{i_k}\}$.
By setting the canonical coordinate system on $\ml T_{i}$ in which 
$\ol x_{i}$ is the origin and $u_{1}^{(i)},\ldots,u_{d}^{(i)}$ are
the directions of the coordinate axes, we can regard $\wt x_{i_j}$, $j=1,\ldots, k$, as vectors in $\mb R^{d}$, whose coordinates are given by the first 
$d$ rows of $\Sa^{(i)}V^{(i)T}$. That is,
\begin{align}\label{coo}
\begin{bmatrix}
\wt x_{i_{1}} & \cdots & \wt x_{i_{k}}
\end{bmatrix}
=\begin{bmatrix}
\sa_{1}^{(i)}\\
 & \ddots\\
 &  & \sa_{d}^{(i)}
\end{bmatrix}
\begin{bmatrix}
v_{1}^{(i)T}\\
\vdots\\
v_{d}^{(i)T}
\end{bmatrix}
=\begin{bmatrix}
\sa_{1}^{(i)}v_{1}^{(i)T}\\
\vdots\\
\sa_{d}^{(i)}v_{d}^{(i)T}
\end{bmatrix}.
\end{align}
The $j$-th row $\sa_{j}^{(i)}v_{j}^{(i)T}$ hence represents the $j$-th 
coordinate function on $\wt{\ml U}_i$.

The geometrical motivation of Step \ref{H2} is that,
if the dimension $d$ of the target space
is the same as that of the data manifold $\ml M$,
then $\ml T_{i}$ is an approximation of the tangent space at $x_{i}$.
Therefore, $\wt{\ml U}_i$ represents the tangential component of $\ml U_i$.
This motivation however is totally ignored in the implementation, and 
the algorithm of HLLE can run without checking whether $d=\dim \ml M$.
Thus, there comes a question: Is this equality in dimension important, or 
can it be safely ignored?
Possibly a bit unexpectedly, it really matters -- for the original HLLE.
Of course, if $d<\dim\ml M$, it is no surprise that there will be a heavy loss in fidelity.
What interesting is that choosing $d>\dim\ml M$ would
also cause HLLE to produce unwanted results.
We will discuss about this issue and give a simple solution to it in Section \ref{sec:tlle}.
For now let us assume $d=\dim\ml M$. 

Now move on to \ref{H3}.
Our definition of $G^{(i)}$ in \ref{H3} follows the original
paper \cite{hlle}, while some researchers (for example \cite{wang2012,ye2015}) 
set $G^{(i)}$ to be $\left[\bs 1_{k}\ F^{(i)}\ F^{(i)}\boxtimes F^{(i)}\right]$,
where $F^{(i)}:=[\sa^{(i)}_1v^{(i)}_1\,\cdots\, \sa^{(i)}_d v^{(i)}_d]\in\mb{R}^{k\times d}$.
In normal situations we have $\mbox{rank}(M^{(i)})\ge d$,
and hence its first $d$ singular values 
$\sa_{1}^{(i)},\ldots,\sa_{d}^{(i)}$
are nonzero. We will assume this throughout the paper. Thus the only difference between the two definitions
of $G^{(i)}$ is that each corresponding column may differ by a nonzero
factor. As a consequence, they give
rise to the same $H^{(i)}$ after performing the Gram-Schmidt process.

In \cite{hlle}, $H^{(i)}$ is claimed to be a Hessian
estimator on $\ml U_i$, and the motivation of asking $\ml Y$ to solve the 
minimization problem in \ref{H4} is based on 
the fact that if $\ml Y$ can be obtained from
$\ml X$ through some isometric mapping from $\ml M$ to $\mb R^d$ 
(the most ideal situation),
then the global intrinsic coordinate functions on $\ml M$ (which are represented
by the rows of $Y$) should have zero second order derivatives 
everywhere. 
These claims themselves are worthy of more explanations and discussions.
However, we shall not pursue this direction here. Instead, we will provide
a much simpler interpretation of the mechanism behind \ref{H3} and \ref{H4}.

First let us go back to \ref{H2}. 
Since the columns of $M^{(i)}$ have mean zero, 
\[
0=M^{(i)}\bs 1_{k}=U^{(i)}\Sa^{(i)}V^{(i)T}\bs 1_{k}.
\]
Then, since $U^{(i)}$ is invertible, 
and since the first $d$ singular values 
$\sa_{1}^{(i)},\ldots,\sa_{d}^{(i)}$ are assumed to be nonzero,
$V_d^{(i)T}\bs 1_k = 0$. 
Thus, the first $1+d$ vectors 
$\bs 1_{k},v_{1}^{(i)},\ldots,v_{d}^{(i)}$ in $G^{(i)}$
are already orthogonal to each other.
As a consequence, the Gram-Schmidt process
in \ref{H3} does not change them except for normalizing 
$\bs 1_{k}$ to $\fc 1{\sqrt{k}}\bs 1_{k}$
(which is also redundant as we only need the result of the 
last $d(d+1)/2$ columns).

Now let us use $h=(h_{1},\ldots,h_{k})^{T}$ to denote any
one of the $d(d+1)/2$ columns of $H^{(i)}$.
By construction $h$ is a unit vector perpendicular to 
$\bs 1_{k},v_{1}^{(i)},\ldots,v_{d}^{(i)}$. In particular, it is perpendicular to
$\sa_1^{(i)} v_1^{(i)},\ldots, \sa_d^{(i)}v_d^{(i)}$.
From \eqref{coo}, this last statement can be written as 
\begin{align*}
\begin{bmatrix}
\wt x_{i_{1}} & \cdots & \wt x_{i_{k}}
\end{bmatrix}
\begin{bmatrix}
h_1 \\\vdots\\ h_k
\end{bmatrix} = 0,
\end{align*}
which is nothing but a linear relation on $\wt{\ml U}_i$:
\begin{equation}
h_{1}\wt x_{i_{1}}+\cdots+h_{k}\wt x_{i_{k}}=0.\tag{h1}\label{eq:h1}
\end{equation}
Thus, the use of $H^{(i)}$ is that each of its columns describes a linear relation 
on $\wt{\ml U}_{i}$. With all of the columns, it then provides 
a total of $d(d+1)/2$ linear relations.
All together they can be expressed as
\begin{equation}
\begin{bmatrix}\wt x_{i_{1}} & \cdots & \wt x_{i_{k}}\end{bmatrix}H^{(i)}=0.\label{eq:hlo}
\end{equation}
On the other hand, the facts that $h$ is perpendicular to $\bs 1_{k}$
and that $h$ is a unit vector amount to two constraints on the
coefficients: 
\begin{align}
 & \sum_{j=1}^k h_{j}=0,\tag{h2}\label{eq:h2}\\
 & \sum_{j=1}^k h_{j}^{2}=1.\tag{h3}\label{eq:h3}
\end{align}
From the above point of view, the purpose of Step \ref{H4} is also clear:
It simply asks $\ml Y$ to satisfy the same local linear relations
as \eqref{eq:hlo} as best as possible.
Thus what HLLE does, in this new interpretation, is really 
in the same vein as LLE, 
only that their ways of describing local linear relations are different
(see Section \ref{sec:llehlle} for some comparisons). 

Now, for convenience let us call any vector $(h_1,\ldots,h_k)$ which satisfies 
\eqref{eq:h1} $\sim$ \eqref{eq:h3} an $h$-weight.
An essential simplification which can be made from the above 
new understanding of HLLE is that we may
choose $H^{(i)}$ arbitrarily
as long as its columns are constituted by $h$-weights!
It needs not be constructed from 
the specially designed $V_{d}^{(i)}\boxtimes V_{d}^{(i)}$ 
in \ref{H3}, and also its number of columns needs not be $d(d+1)/2$.
The number of columns in $H^{(i)}$, call it $m$, just indicates 
how many $h$-weights we would like to use to 
characterize the linear relation on $\wt{\ml U}_i$. 
It is a number we can decide at will, subject only to the
slight restriction $k\ge 1+d+m$ (replacing
the stronger requirement $k\ge 1+d+d(d+1)/2$ given in \ref{H1}),
since the vectors $\bs 1_k,v_1^{(i)},\ldots, v_d^{(i)}$ 
and the columns of $H^{(i)}$ have to be linearly independent. 
In other words, 
\begin{quote}
we have to choose $k\ge d+2$ in \ref{H1}, 
and then $m$ can be any number between $1,2,\ldots,k-d-1$.
\end{quote}

To implement the new idea, it is still convenient to make use of the Gram-Schmidt 
process to generate $h$-weights. The only modification from the 
original HLLE is that we can replace $V_d^{(i)}\boxtimes V_d^{(i)}$ by 
other matrices, as long as the columns of $G^{(i)}$ form a linearly 
independent set. For this purpose, it is reasonable to resort to 
a random construction. Precisely, we replace \ref{H3} by the following:
\begin{itemize}
\item Select $m$ random vectors $r_{1},\ldots,r_{m}$ in $\mb R^{k}$,
and perform the Gram-Schmidt process to the columns of 
\[
G^{(i)}:=\begin{bmatrix}\bs 1_{k} & v_{1}^{(i)} & \cdots & v_{d}^{(i)} & r_{1} & \cdots & r_{m}\end{bmatrix}
\]
to obtain a new matrix with orthonormal columns. Then set 
$H^{(i)}\in\mb R^{k\times m}$ to
be the submatrix consisting of the last $m$ columns. 
\end{itemize}
The new method will be called tangential LLE (TLLE).
We stress again that here we only consider cases with $d=\dim\ml M$.
In Section \ref{sec:tlle} we will add one further modification in
order to cover also cases with $d>\dim\ml M$, and a written down 
algorithm for that final version will be given there.

\begin{figure}
\includegraphics[width=1\textwidth]{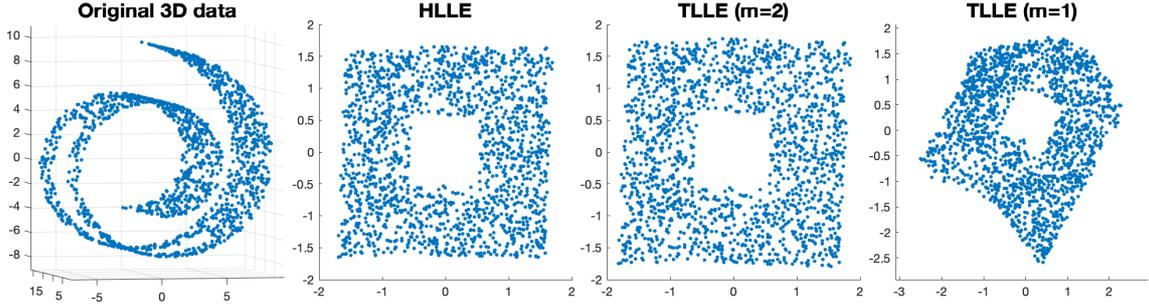}
\caption{\label{fig:ts}Numerical results for Swiss roll with a hole by 
HLLE ($k=8$) and TLLE ($k=8$; $m=2$ and $m=1$).}
\end{figure}

We also stress that $k\ge d+2$ is a minimum restriction in principle. 
From experiments, larger $k$ is usually needed.
On the other hand, 
much fewer $h$-weights than $k-d-1$ may work pretty well. 
Nevertheless, it looks like using multiple weights, i.e. $m\ge 2$, is crucial to ensure good results.
Figure \ref{fig:ts} shows some numerical simulations 
on the Swiss roll with a hole ($d=2$).
Each of the simulations adopts $k=8$. For TLLE, largest possible $m$ is $8-2-1=5$, while 
$m=2$ already gives equally good result as HLLE (which can be regarded as using $m=3$).
In fact, the outcomes of HLLE and TLLE ($m=2$) look identical! 
(The two pictures differ only in size, which is merely 
a consequence of scaling.) 
This is not a coincidence but is usually the case from our repeated experiments. 
Though astonishing at first sight, this phenomenon simply reflects the fact 
that both results are almost perfect: They both unfold
the 3D data faithfully without messing up any part.

\section{Comparison of LLE and TLLE}\label{sec:llehlle}

In this section we also consider $d=\dim\ml M$.
We have mentioned that TLLE (in particular HLLE) implements essentially the 
same idea as LLE: Asking $\ml Y$ to fit the 
local linear relations of $\ml X$ as best as possible.
The main differences lie in their ways of describing local linear relations. 
In this section we take a close look at some of the differences.
Readers who would like to know the practical usage of TLLE quickly
may read only point (A)
below to have a basic understanding of ``projection pattern'', and then go 
straight to the next section.

For convenience we give a review of the procedure of LLE first. 
It goes as follows:
\begin{enumerate}[align=left]
\item[(L1)] For each $i=1,\ldots,N$ let $\ml U_{i}=\{x_{i_{1}},\ldots,x_{i_{k}}\}\subset\ml X\setminus\{x_{i}\}$
be a $k$-nearest neighborhood of $x_{i}$. For simplicity we consider
$k$ to be a fixed number for all $i$.
\item[(L2)] On each $\ml U_i$, solve the minimization problem
\begin{equation}
\underset{(w_{1},\ldots,w_{k})\in\mb R^{k}}{\mbox{argmin}}\|x_{i}-\sum_{j=1}^{k}w_{j}x_{i_{j}}\|^{2}\quad\mbox{s.t.}\quad\sum_{j=1}^{k}w_{j}=1,\tag{P1}\label{eq:p1}
\end{equation}
and let $w^{(i)}=(w_{1}^{(i)},\ldots,w_{k}^{(i)})$ be a solution. 
\item[(L3)] Set $\ml Y$ to be a solution to
\begin{equation*}
\underset{\{y_{i}\}_{i=1}^{N}\subset\mb R^{d}}{\mbox{argmin}}\sum_{i=1}^{N}\|y_{i}-\sum_{j=1}^{k}w_{j}^{(i)}y_{i_{j}}\|^{2}\quad\mbox{s.t.}\quad YY^{T}=I,
\end{equation*}
where $Y=\left[y_{1}\ \cdots\ y_{N}\right]\in\mb R^{d\times N}$.
\end{enumerate}

Recall that we call a vector $h=(h_1,\ldots,h_k)$ that 
satisfies \eqref{eq:h1} $\sim$ \eqref{eq:h3} in the previous section an $h$-weight.
Similarly, let us call a solution
$w=(w_1,\ldots,w_k)$ of Problem \eqref{eq:p1} above a $w$-weight.
It gives rise to a (possibly approximate) linear relation 
of the form
\begin{align}\label{lr}
x_i \approx w_1 x_{i_1}+w_2 x_{i_2}+\cdots +w_k x_{i_k}.
\end{align}
Following are some differences between LLE and TLLE.

\begin{itemize}
\item[(A)] The local linear relations in LLE take the 
full $D$-dimensional data
$\ml U_{i}$ into consideration, while those in TLLE only concern
the ``tangential component'' $\wt{\ml U}_{i}$. 
This is a fundamental difference, which 
protects TLLE from producing a type of unwanted result called 
``projection pattern''. 
Such a result looks like some direct projection of 
$\ml X\subset\mb R^D$ onto a $d$-dimensional subspace.
They are observed in the application of LLE when 
the local linear relations \eqref{lr} for all $i=1,\ldots,N$
are exact or highly approximately true. The core reason is that 
the relations \eqref{lr} are preserved by 
any global (i.e. independent of $i$) linear transformation
from $\mb R^D$ to $\mb R^d$ (see \cite{lin2021avoiding} for 
more details). By contrast, relations of the form 
\eqref{eq:h1} do not have this property:
\begin{align*}
h_1\wt x_{i_1}+\cdots+ h_k\wt x_{i_k}\ \not\Rightarrow\ 
h_1 \wt {Ax}_{i_1}+\cdots + h_k\wt{Ax}_{i_k},
\end{align*}
where $A:\mb R^D\to \mb R^d$ denotes a general linear map indenepdent
of $i$, and $\{\wt{Ax}_{i_1},\ldots$ $,\wt{Ax}_{i_k}\}$ 
denotes the tangential component ($d$-dimensional principal component)
of $\{Ax_{i_1},\ldots,Ax_{i_k}\}$.
\item[(B)] Another important point which makes 
the performance of TLLE much better than that of LLE
is that TLLE can exploit multiple $h$-weights to describe 
linear relations on $\wt{\ml U}_{i}$, while in LLE only one 
linear relation described by $w$-weight is considered for $\ml U_{i}$. 
In fact, the performance of TLLE is evidently worse when using only $m=1$ 
(as Figure \ref{fig:ts} shows). 
Note however that the idea of using multiple weights to improve LLE is not new.
See for example \cite{mlle}.
\item[(C)] In the linear relation \eqref{lr}, $x_{i}$ is something regarded as
the ``center'' of $\ml U_{i}$, despite the fact that it may actually
be far from the center geometrically (for example when $x_{i}$ is
a point close to the boundary of $\ml M$). On the other hand, in the relation 
\eqref{eq:h1} there is no point which is regarded as playing the central role. 
Nevertheless, this difference is somewhat superficial and 
might be of less significance.
For example, if say $h_{1}\ne 0$ in \eqref{eq:h1}, then by dividing
\eqref{eq:h1} and \eqref{eq:h2} by $h_{1}$ (and abandoning the
size-controlling constraint \eqref{eq:h3}), we get 
\begin{align*}
 & \wt x_{i_{1}}=\wt w_{2}\wt x_{i_{2}}+\cdots+\wt w_{k}\wt x_{i_{k}}\\
 & \wt w_{2}+\cdots+\wt w_{k}=1,
\end{align*}
where $\wt w_{j}:=-h_{j}/h_{1}$. Thus we get a linear relation of
LLE type, with $\wt x_{i_{1}}$ playing the role of the center. Similarly, it is also easy 
to rewrite an LLE-type linear relation into that of 
a TLLE-type by multiplying a suitable factor.
%given a (possibly approximate) linear relation of LLE type:
%\[
%x_{i}\approx\sum_{j=1}^{k}w_{j}x_{i_{j}},\quad\mbox{with}\quad\sum_{j=1}^{k}w_{j}=1.
%\]
%We can transform it into a ``decentralized'' form: 
%\begin{align*}
% & \wt h_{0}x_{i}+\wt h_{1}x_{i_{1}}+\cdots+\wt h_{k}x_{i_{k}}\approx0,\\
% & \sum_{j=0}^{k}\wt h_{j}=0,\quad\sum_{j=0}^{k}\wt h_{j}^{2}=1,
%\end{align*}
%where $\wt h_{0}=(1+\sum_{j=1}^{k}w_{j}^{2})^{-1/2}$, and $\wt h_{j}=-w_{j}(1+\sum_{j=1}^{k}w_{j}^{2})^{-1/2}$
%for $j=1,\ldots,k$.
\end{itemize}

Before ending this section, we would like to share one more interesting 
observation about the selection of $k$-nearest neighborhood. 
In LLE, the reason we exclude $x_{i}$ from $\ml U_{i}$ is clear,
otherwise we may obtain the trivial linear relation $x_{i}=x_{i}$
from Problem \eqref{eq:p1}. This practice is inherited by HLLE.
However, a casual examination of 
HLLE or TLLE reveals no harm to include $x_{i}$ in $\ml U_{i}$.
For example, we may 
set $x_{i_{1}}$ to be $x_{i}$, and $x_{i_{2}},\ldots,x_{i_{k}}$
are $k-1$ nearest points to $x_{i}$. 
However, in doing so a subtle problem occurs:
we may have 
$\ml U_{i}=\ml U_{j}$ for some $i\ne j$.
Indeed, we observed that such an equality occurs with high probability and 
in abundance, which causes a lot of redundancy since the information
in $H^{(i)}$ and $H^{(j)}$ may overlap. As a consequence, it is 
still a good practice to exclude $x_{i}$ from $\ml U_{i}$ in TLLE. 

\section{A modification for cases with $d>\dim\ml M$}
\label{sec:tlle}

For convenience let us use $d_{\ml M}$ to denote the dimension of
$\ml M$ in this section, and $d$ is reserved for the dimension
of the target space. As is mentioned, in the original idea of HLLE
$d$ is supposed to be the same as $d_{\ml M}$, which however is 
ignored in the algorithm. Indeed, it would
be better to have no such a restriction, since in general a manifold
with nontrivial curvature cannot be well embedded in $\mb R^{d}$
for $d\le d_{\ml M}$, and selecting a larger $d$ might sometimes be more suitable. 
Of course,
this perspective is based on the wish to preserve as much metric information
of the original data $\ml X$ as possible. 
In another direction, when a definitely low dimensional target
space is strongly preferred such as for visualization of data in $\mb R^{2}$
or $\mb R^{3}$, one may even consider $d<d_{\ml M}$. In this case 
substantial loss in fidelity is unavoidable. We will not
discuss this direction.

Back to the assumption $d=d_{\ml M}$. Now the question is:
Does it really matter? Unfortunately, the answer is yes. 
In fact, a naive application of HLLE with $d>d_{\ml M}$ may result in 
``projection patterns'' as LLE (see point (A) in Section \ref{sec:llehlle}).
Numerical illustrations will be given in the end of this section.
The reason for getting these unwanted results should be that, 
since each $\ml U_i$ is already close to a $d_{\ml M}$-dimensional piece, 
its $d$-dimensional principal component 
contains almost the full $D$-dimensional information.
In other words, a linear relation \eqref{eq:h1} for $\wt{x}_{i_1},\ldots,\wt{x}_{i_k}$ 
is nearly an exact linear relation for $x_{i_1},\ldots,x_{i_k}$, 
and hence is almost preserved by any linear map 
from $\mb R^D$ to $\mb R^d$. 
For LLE, it is then recommended in \cite{lin2021avoiding} that one uses 
regularization to disturb the linear relations so that they will not be ``too exact''. 
Here for HLLE, we have another simple solution:
adhere to the rule of ``fitting only the tangent part''.
That is, when $d_{\ml M}<d$, we should fit 
linear relations of the $d_{\ml M}$-dimensional
principal component, instead of the $d$-dimensional one. 

Specifically, we set $\wt{\ml U}_{i}=\{\wt x_{i_{1}},\ldots,\wt x_{i_{k}}\}$
to be the $d_{\ml M}$-dimensional principal component of $\ml U_{i}$.
Accordingly, an $h$-weight for $\wt{\ml U}_{i}$ is a unit vector
in $\mb R^{k}$ which is perpendicular to the $d_{\ml M}+1$ vectors
$\bs 1_{k},v_{1}^{(i)},\ldots,v_{d_{\ml M}}^{(i)}$.
The restriction of the number $k$ then becomes 
$k\ge d_{\ml M}+2$, which allows at most $k-d_{\ml M}-1$ 
linearly independent $h$-weights for each $\wt{\ml U}_{i}$. 
On the other hand, since we still want to find a representation
$\ml Y$ of $\ml X$ in $\mb R^{d}$, the minimization problem
in \ref{H4} is unchanged. 
We summarize the algorithm for this final version of TLLE below. 

\begin{mdframed}
\begin{center}
TLLE Algorithm
\end{center}\medskip

\noindent\textbf{Inputs:} (1) $\ml X=\{x_{i}\}_{i=1}^{N}\subset\mb R^{D}$, (2) two
positive integers $d_{\ml M}$ and $d$, with $1\le d_{\ml M}\le d<D$,
(3) a positive integer $k\ge d_{\ml M}+2$, and (4) a positive integer
$m\le k-d_{\ml M}-1$.\smallskip

\noindent\textbf{Output:} $\ml Y=\{y_{i}\}_{i=1}^{N}\subset\mb R^{d}$.\smallskip

\noindent\textbf{Procedure:}
\begin{enumerate}%[align=left]
\item For each $i=1,\ldots,N$ let 
$\ml U_{i}=\{x_{i_{1}},\ldots,x_{i_{k}}\}\subset\ml X\setminus\{x_{i}\}$
be a $k$-nearest neighborhood of $x_{i}$.
\item Let $M^{(i)}=\left[x_{i_{1}}-\ol x_{i}\ \cdots\ x_{i_{k}}-\ol x_{i}\right]\in\mb{R}^{D\times k}$,
where $\ol x_{i}=\fc 1k\sum_{j=1}^{k}x_{i_{j}}$, and compute its
singular value decomposition:
\begin{align*}
M^{(i)}=U^{(i)}\Sa^{(i)}V^{(i)T}.
\end{align*}
Let $v_{1}^{(i)},\ldots,v_{d_{\ml M}}^{(i)}$ be the first $d_{\ml M}$
columns of $V^{(i)}$.
\item Select $m$ random vectors $r_{1},\ldots,r_{m}$ in $\mb R^{k}$,
and perform the Gram-Schmidt process to the columns of the matrix
\[
\begin{bmatrix}\bs 1_{k} & v_{1}^{(i)} & \cdots & v_{d_{\ml M}}^{(i)} & r_{1} & \cdots & r_{m}\end{bmatrix}
\]
to obtain a new matrix. Then set $H^{(i)}\in\mb R^{k\times m}$ to
be the submatrix of this new matrix consisting of the last $m$ columns. 
\item Set $\ml Y$ to be a solution to the minimization problem
\begin{equation*}
\underset{\{y_{i}\}_{i=1}^{N}\subset\mb R^{d}}{\mbox{argmin}}\sum_{i=1}^{N}\|Y_{i}H^{(i)}\|^{2}\quad\mbox{s.t.}\quad YY^{T}=I,%\label{eq:hmm}
\end{equation*}
where $Y=\left[y_{1}\ \cdots\ y_{N}\right]\in\mb R^{d\times N}$ and
$Y_{i}=\left[y_{i_{1}}\ \cdots\ y_{i_{k}}\right]\in\mb R^{d\times k}$.
Note that the norm $\|\cdot\|$ for matrices denotes the Frobenius norm.
\end{enumerate}
\end{mdframed}

Some remarks about the practical implementation are in order. 

\begin{rem}
Here we recall the routine way to solve the minimization problem 
in Step 4. First, rewrite the summation
$\sum_{i=1}^{N}\|Y_{i}H^{(i)}\|^{2}$ into a single $\|YH\|^2$, where $H$ is 
an $N\times mN$ matrix defined as follows:
For $i=1,\ldots,N$, its submatrix of the $((i-1)m+1)$-th, $((i-1)m+2)$-th,..., $((i-1)m+m)$-th columns 
and the $i_1$-th, $i_2$-th,..., $i_k$-th rows is $H^{(i)}$; other entries are all zero. 
Then any minimizer $\{y_i\}_{i=1}^N$ is given by 
\begin{align*}
\begin{bmatrix}
y_1 & \cdots & y_N
\end{bmatrix}^T = \begin{bmatrix}
g_1 & \cdots & g_d
\end{bmatrix},
\end{align*}
where $g_1,\ldots,g_d$ are orthonormal eigenvectors of $HH^T$ corresponding
to the smallest $d$ eigenvalues (counting multiplicity). We assume that their 
corresponding eigenvalues are already arranged in ascending order.  
However, in this way $g_1$ is supposed to be $\frac{1}{\sqrt{N}}\bs 1_N$,
an eigenvector of $HH^T$ corresponding to the smallest eigenvalue $0$.
This eigenvector is redundant for our purpose of embedding $\ml X$ into 
$\mb R^d$, and can be omitted. 
Alternatively, this omission can be regarded as adding one further constraint
$Y\bs 1_N = 0$ to the minimization problem, meaning that we ask 
$\{y_i\}_{i=1}^N$ to have mean zero. 
Therefore, what is really executed in the algorithm 
is setting $g_1,\ldots, g_d$ to be the 2nd to the $(d+1)$-th eigenvalues. 
As discussed in \cite{lin2021avoiding} (for LLE), this practice may be problematic 
if there are more than one linearly independent eigenvectors corresponding to the eigenvalue
$0$.
For TLLE, such a problem would almost never occur
as long as multiple weights is adopted. Since $H$ has size $N\times mN$
and is constructed randomly, $\bs 1_N$ is safely the only eigenvector of $HH^T$
corresponding to $0$ if $m\ge 2$. 
\end{rem}

\begin{rem}
In applications when the dimension of $\ml M$ is 
not well understood in advance, $d_{\ml M}$ can be realized 
as the number of significant singular values in $\Sa^{(i)}$. 
Although this criterion sounds somewhat imprecise, 
from our experience it is usually easy to make a 
judgment, at least for artificial examples. For example, a typical
neighborhood $\ml U_{i}$ for the Swiss roll may have $\sa_{1}^{(i)}=3.1475$,
$\sa_{2}^{(i)}=2.4907$ and $\sa_{3}^{(i)}=0.2346$. It is obvious that the first two 
are significant and the third one is not, showing that the data manifold is  
two-dimensional. 
We may look at this issue from another angle: If $\ml X$ is such that 
there shows no definite ``number of significant singular values''
from principal component analysis on the neighborhoods $\ml U_i$, 
then the ``manifold perspective'' on $\ml X$ would be questionable, 
and applying TLLE to it may not produce a reliable result. 
\end{rem}

\begin{figure}
\includegraphics[width=1\textwidth]{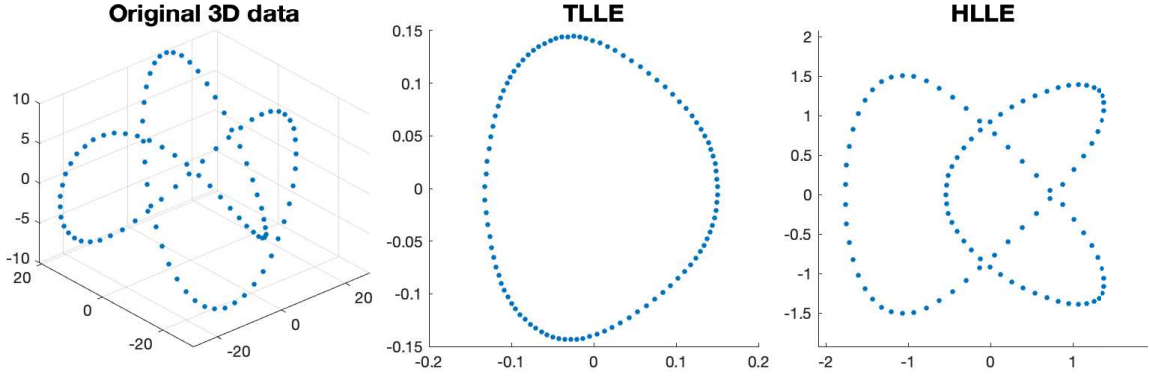}
\caption{A comparison between TLLE and HLLE on the Trefoil knot.\label{fig:tref}}
\end{figure}

Finally, we would like to give some numerical examples to demonstrate 
how TLLE can avoid projection patterns while HLLE can not. 
For this purpose, we have to consider three different dimensions: $d_{\ml M}<d<D$. 
The simplest case that is visualizable is $d_{\ml M}=1$, $d=2$, and $D=3$. 
This means that the data manifold is a space curve, and we are to find a 
representation for it on the plane. Figure \ref{fig:tref} shows numerical results for 
applying TLLE and HLLE to the so-called
``Trefoil knot''. 
Note that in this case, although the dimension of the manifold is one, 
it can not be embedded in the real line. 
We see that TLLE realizes the knot as a topological circle, 
while HLLE projects (possibly with some distortion) the knot onto the plane, causing self-intersections. 

\begin{figure}
\begin{center}
\includegraphics[width=0.9\textwidth]{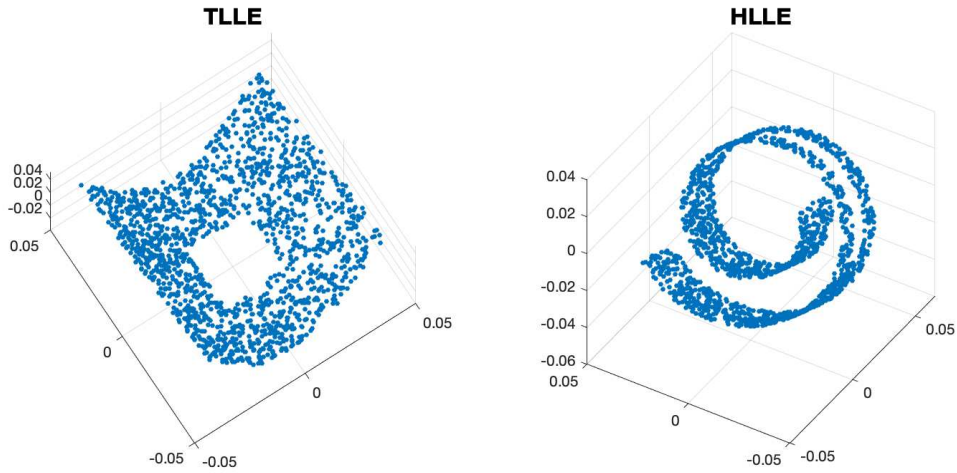}
\end{center}
\caption{Results of TLLE and HLLE on the Swiss roll with a hole that is isometrically 
embedded in $\mb{R}^9$, and then mapped back
to $\mb{R}^3$.\label{fig:hsr}}
\end{figure}

Figure \ref{fig:hsr} shows another illustrative example. 
It is obtained by first performing an isometrical embedding 
of the Swiss roll with a hole in $\mb{R}^9$, and then -- 
suppose we do not know that it can be well unfolded on the plane --
applying TLLE and HLLE to map it back to $\mb R^3$.
From the knowledge of $d_{\ml M}=2$, TLLE is capable of identifying the intrinsic 
geometry, and unfolds the Swiss roll to some extent in $\mb{R}^3$. 
On the other hand, HLLE just performs some projection from 
$\mb{R}^9$ to $\mb{R}^3$. Usually the results are similar to the original 
3D data and look not bad, while in a few cases they happen to be
highly compressed along some direction and are not satisfactory.

\section{Conclusion}

In this paper we explained that HLLE can be viewed as implementing the same
idea as LLE: Asking the dimension-reduced data $\{y_i\}_{i=1}^N\subset\mb R^d$ 
to satisfy the same local linear relations as the original data 
$\{x_i\}_{i=1}^N\subset\mb R^D$ 
as best as possible. The main differences lie in the following two 
points: 
\begin{itemize}
\item[(A)]HLLE only considers linear relations of the $d$-dimensional 
principal component for each neighborhood;
\item[(B)]HLLE exploits multiple (originally $d(d+1)/2$) linear relations
for each neighborhood.
\end{itemize}
With these being clear, we proposed a simplification where
the ``Hessian estimator'' was replaced by 
randomly constructed weight matrices. 
Moreover, when $d$ is greater than $d_\ml M$ (the dimension of the 
data manifold), HLLE may produce projection-like results. To avoid
this problem we suggested that the $d$-dimensional principal 
component in (A) should be 
replaced by the $d_\ml M$-dimensional one, which represents
the tangential component of the original data. 
Combining the above simplification and modification, we finally achieved a
new LLE-type method which was named tangential locally linear embedding (TLLE).
It is simpler and more robust than HLLE.

So far, our numerical experiments are focused on artificial datasets 
such as the Swiss roll with a hole, and the performances of TLLE look excellent. 
Whether it is also helpful 
in producing reliable results for real world data or data with noise 
is an important direction for future investigation. 

\section*{Acknowledgment}
This work is supported by Ministry of Science 
and Technology of Taiwan under grant number MOST110-2636-M-110-005-. 
The authors thank Dr. Hau-Tieng Wu for reading and giving valuable comments on the first version of our manuscript.

\bibliographystyle{plain}
\bibliography{tlle}

\end{document}